\documentclass[lettersize,journal]{IEEEtran}

\usepackage{amsmath,amssymb,amsfonts}
\usepackage{mathtools}
\usepackage{algorithmic}
\usepackage{graphicx}
\usepackage{booktabs}
\usepackage{textcomp}
\usepackage{xcolor}
\usepackage{xspace}
\usepackage{multicol}
\usepackage{multirow}
\usepackage{url}
\usepackage{listings}
\usepackage[utf8]{inputenc}
\usepackage{bm}
\usepackage{subcaption}
\usepackage{gensymb}
\usepackage{makecell}
\usepackage{etoolbox}
\usepackage{hyperref}
\usepackage{soul}
\usepackage{tikz}
\usepackage[ruled,vlined]{algorithm2e}
\usepackage{pgf-umlsd}
\usepackage{balance}
\usepackage{float}
\usepackage{soul}
\usepackage{bigstrut}

\DeclareMathOperator*{\argmin}{arg\,min}

\newcommand{\ignore}[1]{}
\usepackage{xspace}
\usepackage{ifthen}
\newboolean{showcomments}
\setboolean{showcomments}{true} 
\ifthenelse{\boolean{showcomments}}{
\newcommand{\mynote}[3]{\noindent{\color{#3}\textbf{#1\xspace} #2}}}
{ \newcommand{\mynote}[3]{}
}
\definecolor{mypurple}{rgb}{0.6,0.4,0.8}
\definecolor{mygreen}{rgb}{0.31,0.61,0.12}

\definecolor{myblue}{RGB}{0, 114, 178} 
\definecolor{myorange}{RGB}{230, 159, 0} 
\definecolor{mygreen}{RGB}{0, 158, 115} 
\definecolor{mygrey}{RGB}{153, 153, 153} 

\DeclareRobustCommand\blueline
{\tikz[baseline=0.0ex]\draw[myblue, line width=1pt] (0,0.1) -- (0.35,0.1);}
\DeclareRobustCommand\orangeline
{\tikz[baseline=0.0ex]\draw[myorange, line width=1pt] (0,0.1) -- (0.35,0.1);}
\DeclareRobustCommand\redline
{\tikz[baseline=0.0ex]\draw[red, line width=1pt] (0,0.1) -- (0.35,0.1);}

\DeclareRobustCommand\blackline
{\tikz[baseline=0.0ex]\draw[black, line width=1pt] (0,0.1) -- (0.35,0.1);}

\DeclareRobustCommand\greydashedline
{\tikz[baseline=0.0ex]\draw[mygrey, dashed, line width=1.2pt] (0,0.1) -- (0.35,0.1);}

\newcommand{\ie}{\textit{i.e.,}\xspace}
\newcommand{\eg}{\textit{e.g.,}\xspace}
\newcommand{\wrt}{\textit{w.r.t.}\xspace}
\newcommand{\aka}{\textit{a.k.a.}\xspace}

\newtheorem{definition}{\textbf{Definition}}

\newcommand{\mysubsubsection}[1]{\vspace{2pt} \noindent \textbf{#1}}

\newif\iftex
\textrue

\begin{document}

\title{Quantifying and Localizing Usable Information Leakage from Neural Network Gradients}

\author{
\IEEEauthorblockN{Fan Mo\IEEEauthorrefmark{1},
Anastasia Borovykh\IEEEauthorrefmark{2}, 
Mohammad Malekzadeh\IEEEauthorrefmark{1}, 
Soteris Demetriou\IEEEauthorrefmark{1},
Deniz Gündüz\IEEEauthorrefmark{1},
Hamed Haddadi\IEEEauthorrefmark{1}\\}
\IEEEauthorblockA{\IEEEauthorrefmark{1}Imperial College London}
\IEEEauthorblockA{\IEEEauthorrefmark{2}University of Warwick}
}

\maketitle

\begin{abstract}
In collaborative learning, clients keep their data private and communicate only the computed {\em gradients} of the deep neural network being trained on their local data. Several recent attacks show that one can still extract private information from the shared  network's gradients compromising clients' privacy.
In this paper, to quantify the {\em private information leakage from gradients} we adopt {\em usable information theory}. We focus on two types of private information: {\em original} information in {\em data reconstruction attacks} and {\em latent} information in {\em attribute inference attacks}. Furthermore, a {\em sensitivity analysis} over the gradients is performed to explore the underlying cause of information leakage and validate the results of the proposed framework. Finally, we conduct numerical evaluations on six benchmark datasets and four well-known deep models. We measure the impact of training hyperparameters, \eg~batches and epochs, as well as potential defense mechanisms, \eg~dropout and differential privacy.
Our proposed framework enables clients to localize and quantify the private information leakage in a layer-wise manner, and enables a better understanding of the sources of information leakage in collaborative learning, which can be used by future studies to benchmark new attacks and defense mechanisms. 
\end{abstract}

\section{Introduction}\label{sec:intro}

Collaborative learning~\cite{mcmahan2017communication, yang2019federated} allows private data owners 
(holding personal, medical, financial, or governmental datasets~\cite{hard2018federated, rieke2020future, zheng2022applications})
to participate in training a deep neural network~(DNN) model for some {\em target} task. Instead of sharing their private datasets, {\em clients} only share the {\em gradients} (or updated parameters) of the locally trained model at each iteration. The motivation of collaborative learning is to utilize local computational power while complying with data protection regulations~\cite{EUdataregulations2018}. However, as noted by existing {\em attacks} to collaborative learning~\cite{zhu2019deep, zhao2020idlg, melis2019exploiting, geiping2020inverting, nasr2019comprehensive}, the shared gradients might be enough for either the central server or other clients to infer private {\em attributes} of a data owner or even to {\em reconstruct} their private data. These ever-increasing attacks on the shared gradients call into doubt the privacy promised by collaborative learning. Yet, it is not fundamentally understood which {\em layers} of a DNN can facilitate leaking private information more, and to what {\em extent} different types of information are stored in the shared gradients of each layer. 

Despite many empirical privacy attacks on certain applications of collaborative learning, only a few works have recently started preliminary investigations on the performance of these attacks and the possibility of defending against them (see \eg \cite{huang2021evaluating, liu2022ml, usynin2022zen}). The work of \cite{liu2022ml} further developed a Python-based software that can be reused partially for conducting and measuring attacks. 
Based on these previous works, there still exist several open questions.
i) Available works, proposing privacy measurements, mainly focus on experimental results. Few recent works~\cite{zhu2020r, fan2020rethinking} explain why data reconstruction attacks can be successful in certain applications, 
but we still lack a {\em unifying theory} for measuring different types of information contained in gradients.
ii) Previous investigations are coarse-grained and look at the DNN model as a whole, which means that they do not {\em localize} which layers in a target DNN contain more private information for specific attacks. 
iii) Previous approaches only present the results of applying existing attacks and defenses, and do not allow us to understand more generally the {\em underlying reasons} of how information leakage can be linked to other properties of DNN gradients; \eg~the sensitivity to the changes in the input data.

In a separate line of work, information theory is used for measuring the amount of original or latent information that is captured in the outputs of different DNN layers and how this information evolves during the training procedure~\cite{tishby2000information, goldfeld2019estimating, shwartz2017opening, saxe2019information, achille2019information}. To understand the information flow through DNN layers in the forward pass, Shannon's mutual information (MI)~\cite{shwartz2017opening, saxe2019information, goldfeld2019estimating} between each layer's intermediate representations and either the input data (\ie the original info) or the output label (\ie the public latent info) is estimated during each iteration. Nevertheless, it is well-known that the estimation of Shannon's entropy and MI for high-dimensional data can have a high bias or variance~\cite{paninski2003estimation, belghazi2018mutual}. Finally, we remark on the line of work that introduces more general information-theoretical quantities for measuring information leakage, see e.g.  \cite{issa2019operational,liao2019tunable,Rassouli:TIFS:20}. This work however does not quantify specific types of leakage in DNNs. Furthermore, the above-mentioned estimates are difficult to adapt to gradients produced in backpropagation, where analyses based on the data processing inequality (DPI) are not applicable anymore~(see section~\ref{sec:MI} for further explanation).

“In this work, we bridge the gap between the aforementioned approaches, and adopt the concept of {\em usable information~(\aka predictive \mbox{$\mathcal{V}$-information}}~\cite{xu2020theory}) 
to approximate {\em the probability of a successful attack} using a particular family of attack models (to be defined later), given that the attacker is able to obtain a subset of the gradients. We consider two types of private information: i) {\em original information}, the observed data in training datasets, and ii) {\em latent information}, \ie attributes of the observed data; in this way we cover most existing, critical attacks based on gradients. Our proposed measure is based on the probability of recovering certain information (for attribute inference attacks), or of recovering input data within a certain similarity (for reconstruction attacks). To enable our measure to be generalized to private information of any granularity/dimensions, we consider the outcomes of every attack as a probability distribution over the {\em attacker's objective} (to be made precise later), and we approximate the density of this probability distribution.
Hence, we can quantify and localize private information in a layer-wise manner to better understand the ``private information flow'' through gradients of a DNN's layers. To validate our proposed probabilistic measure and to further explore the underlying cause of information leakage, we perform {\em sensitivity analysis} over the gradients with respect to~(\wrt) the two types of private information. This approach is inspired by the concept that sensitivity of gradients \wrt~input/output can reflect model robustness~\cite{novak2018sensitivity}, and robustness, in turn, is related to privacy risks~\cite{han2021understanding}. As sensitivity computation relies only on the trained model, it helps to understand leakages independently of the attack model.

\mysubsubsection{Contributions.} Our paper contributes to privacy-preserving machine learning in the following ways:

\textbf{First}, we introduce a novel explanation of private information leakage using the concept of {\em usable information} theory. The proposed measure allows us to quantify the privacy risks of any selected subset of the shared gradients through a probabilistic view of the attacker's objective. Due to the nature of usable information theory, our framework enables us to differentiate attackers with different computational power,
and thus better understand how much latent and original information leakage can occur. New attacks proposed in the future can be easily integrated into our framework.

\textbf{Second}, utilizing the proposed framework, we perform a {\em layer-wise} localization of private information flow in the backpropagation process of DNNs. Our characterizations reveal that fully-connected layers in benchmark DNN classifiers usually contain the highest latent and original information. We believe such understanding can help to achieve better data protection in environments where one can choose to hide or protect a subset of the DNN layers, \eg~split learning~\cite{thapa2020splitfed}, differential privacy (DP)~\cite{mcmahan2018general}, or trusted execution environments~\cite{mo2020darknetz}.

\textbf{Third}, using a heuristic, general-purpose measure named {\em gradient sensitivity}, we examine the ``predictability'' of information leakage. The sensitivity of the gradient \wrt~the private input is computed via the Jacobian matrix of the gradients. This gives us an additional tool to validate our proposed measure of usable information in an attack-independent manner. 

\textbf{Fourth}, we extensively evaluate the effect of training hyperparameters and defense mechanisms on the probability of information leakage. We show that more {\em epochs} do not significantly reduce the amount of information leakage, but gradient {\em aggregation} can significantly reduce it. We also show that applying dropout and differential privacy on more sensitive layers can achieve better performance in alleviating information leakage.

\section{Background and Related Works}
\begin{table}[t!]
    \caption{\textbf{Notations}. We use lower-case italic (\eg~$x$), lower-case bold (\eg~$\bm{x}$), and upper-case bold (\eg~$\bm{X}$) for deterministic scalars, vectors, and matrices, respectively. Roman-type upper-case (\eg~$\mathrm{X}$) and lower-case (\eg~$\mathrm{x}$) denote random variables (of any dimensions) and their instances, respectively.} 
    \label{tab:notation}
    \footnotesize
    \setlength\tabcolsep{2pt}
    \renewcommand{\arraystretch}{1.1}
    \begin{tabular}{l  p{6.7cm}}
         \multicolumn{1}{l}{\textbf{Symbol}} & \multicolumn{1}{l}{\textbf{Description}}\\\toprule
         $\bm{X}$, $\mathrm{X}$ and $\mathrm{x}$& Original input data or original information \\
         $\bm{G}$, $\mathrm{G}$ and $\mathrm{g}$ & Gradient matrix or information as random variable\\
         $\bm{y}$, $\mathrm{Y}$ and $\mathrm{y}$& Label of input data or information as random variable \\
         $\bm{p}$, $\mathrm{P}$ and $\mathrm{p}$ & Latent attribute or latent information\\
         $\mathrm{O}$ and $\mathrm{o}$ & Attack objective as random variables\\\midrule
        $I(\mathrm{X};\mathrm{G})$  &   Shannon's mutual information between $\mathrm{X}$ and $\mathrm{G}$    \\
        $f \in \mathcal{V}$ & function $f$ in predictive family $\mathcal{V}$ of one attacker \\
        $f[\mathrm{g}](\mathrm{p})$   &  Probability of achieving attack objective $\mathrm{p}$ with $f$ on $\mathrm{g}$ \\
        \scalebox{0.85}[1]{$\hat{I}_{\mathcal{V}}(\mathrm{G}\rightarrow \mathrm{P})$}   &   Empirical usable information from $\mathrm{G}$ to $\mathrm{P}$ under $\mathcal{V}$ \\
        $\mathbf{J}_l^{(\bm{G})}(\bm{X})$   & Input-gradient Jacobian of layer $l$ \\
        $\mathcal{R}_l^{(\bm{X})}$, $\mathcal{R}_l^{(\bm{p})}$  &   Original and latent information risk of layer $l$  \\
        $\mathbb{G}_l^{(\text{0})}$, $\mathbb{G}_l^{(\text{1})}$ & Linear subspaces of $l$'s gradients w/  and w/o an attribute \\
        \scalebox{0.85}[1]{$d_{\mathrm{Gr}} (\mathbb{G}_l^{(\text{0})},\mathbb{G}_l^{(\text{1})})$} & Grassmann distance between subspaces \\\bottomrule
     \end{tabular}
     \vspace{-.2cm}
\end{table}

\subsection{Gradient sharing}
In collaborative machine learning, multiple clients train a common model on their local private datasets~\cite{mcmahan2017communication, kairouz2019advances}. 
The {\em gradients} computed over {\em one batch} of a client's data is denoted by $\bm{G}^1 = \frac{\partial \ell(\bm{X},\bm{y},\bm{W}^0)}{\partial \bm{W}^0}$, where $\bm{X}$ and $\bm{y}$ denote the local training data and its corresponding labels, $\bm{W}^0$ refers to the current model parameters, and $\ell(\cdot)$ refers to the training loss function.
$\bm{G}^1$ consists of the gradients of all layers \wrt the training data; which in typical collaborative learning settings is what is sent to other clients or a central server. 
Conventionally, this process is called \texttt{FedSGD}~\cite{mcmahan2017communication}. To reduce the communication load, it is preferable to consecutively update the received model on \emph{multiple batches} of local data before sharing, called \texttt{FedAvg}~\cite{mcmahan2017communication}. After updating the model on the $t$-th batch, the updated parameters are $\bm{W}^t = \bm{W}^0 + \sum^t_{i=1} \bm{G}^i$, where $\bm{G}^i$ is the gradients of $\bm{W}^i$ computed based on the $i$-th batch of sampled data. When it is not necessary, we might remove the superscripts and refer $\sum^t_{i=1} \bm{G}^i$ as $\bm{G}$ and $\bm{W}^t$ as $\bm{W}$, for simplicity.
Although commonly a client shares the last snapshot of the updated model that consists of the original model and computed gradients (\ie~$\bm{W}^{0} + \bm{G}$), this practically amounts to \emph{sharing} $\bm{G}$ since the central server,  and hence, all the participating clients, know $\bm{W}^{0}$. 
Private information can then be compromised by the attacker as this $\bm{G}$ could contain original private information~($\bm{X}$) or some latent attribute information of it~($\bm{p}$). 
The gradient sharing can be among clients or through a central server; depending on the specific settings~\cite{mo2021ppfl, zhu2019deep,kairouz2019advances}. Gradients shared by the central server are usually aggregated across the clients, meaning that disclosing the private information of a specific client from the aggregated gradients becomes harder. 
Our analysis includes both threat models, as fundamentally disclosing information from gradients at the central server or clients only differs in the level of gradient aggregation.

\subsection{Attack and defense measurements}\label{sec:related_measures}

\begin{table*}[th!]
  \centering
  \caption{Previous works on the measurements of attacks and defenses on neural network gradients.}
    \begin{tabular}{p{2em}p{1em}p{15em}p{40em}}
    \textbf{Attack}    & \textbf{Ref.}   & \textbf{Defenses} & \textbf{Evaluation and Results} \bigstrut\\
    \toprule
   \multirow{5}{*}{DRA} & \multirow{2}{*}{\cite{huang2021evaluating}}   & Gradient pruning, MixUp, InstaHide & Discovers two strong assumptions that current DRAs implicitly make; summarizes defenses and their combinations on defending against DRAs (\eg computation cost) \bigstrut\\
    \cline{2-4}
    & \multirow{3}{*}{\cite{usynin2022zen}} & DP, width and depth, number of clients, batch size, dataset complexity (coarsely) &  Generative encoder and Deep Leakage from Gradients (DLG) attacks for DRA; SSIM and PSNR as metrics for images; showing that model hyperparameters have significant influences on existing attacks \bigstrut\\
    \midrule
      \multirow{3}{*}{AIA} & \multirow{3}{*}{\cite{liu2022ml}} & DP, knowledge distillation, dataset complexity, overfitting, more epochs & Black-box and white-box access using shadow or partial dataset for AIAs; DP-SGD and Knowledge Distillation can only mitigate some of the inference attacks; a modular re-usable software named ML-Doctor \bigstrut\\
    \bottomrule
    \end{tabular}%
  \label{tab:realted_work_measures}%
\end{table*}%

Privacy attacks on collaborative learning are usually conducted on the exposed elements such as the gradients~\cite{melis2019exploiting,zhu2019deep} and intermediate representations~\cite{wang2019beyond}. The attacker can even actively participate in the training process~\cite{he2019model}. Here we focus on the gradient-based attacks that have been recently studied; attacks that are easier to conduct but harder to detect. These attacks are usually based on training some generative adversarial networks and optimizing an attack objective function~\cite{hitaj2017deep,zhu2019deep}. These gradient-based attacks are categorized into: i) the original information, to reconstruct the input data~\cite{zhu2020r,geiping2020inverting,yin2021see}, namely {\em data reconstruction attack}~(DRA), and ii) the latent information, to infer attributes of the input data~\cite{melis2019exploiting}, namely {\em attribute inference attack}~(AIA). 

Along with the rising concern of sharing gradients, studies have also started to investigate the performance of current defenses against these attacks. Table~\ref{tab:realted_work_measures} summarizes the works that are most related to this paper. 
In addition to the measurement of (semi-standard) defense techniques such as DP and dropout, \cite{usynin2022zen} suggests applying simple model adaptation such as increasing the models' depth and the number of clients, which drop the attack success rates in general. 
The authors of~\cite{liu2022ml} further develop a Python-based software ML-doctor that can be reused partially for conducting attacks and measuring the effectiveness of some defense mechanisms. However, these works have several key shortcomings (as explained in Section~\ref{sec:intro}); mostly coarse-grained analysis, purely empirical evaluation, and no grounded justifications for private information leakages.

\subsection{Information flow via neural networks}
\label{sec:MI}
\begin{figure}[t!]
\centering
        \includegraphics[width=0.85\columnwidth]{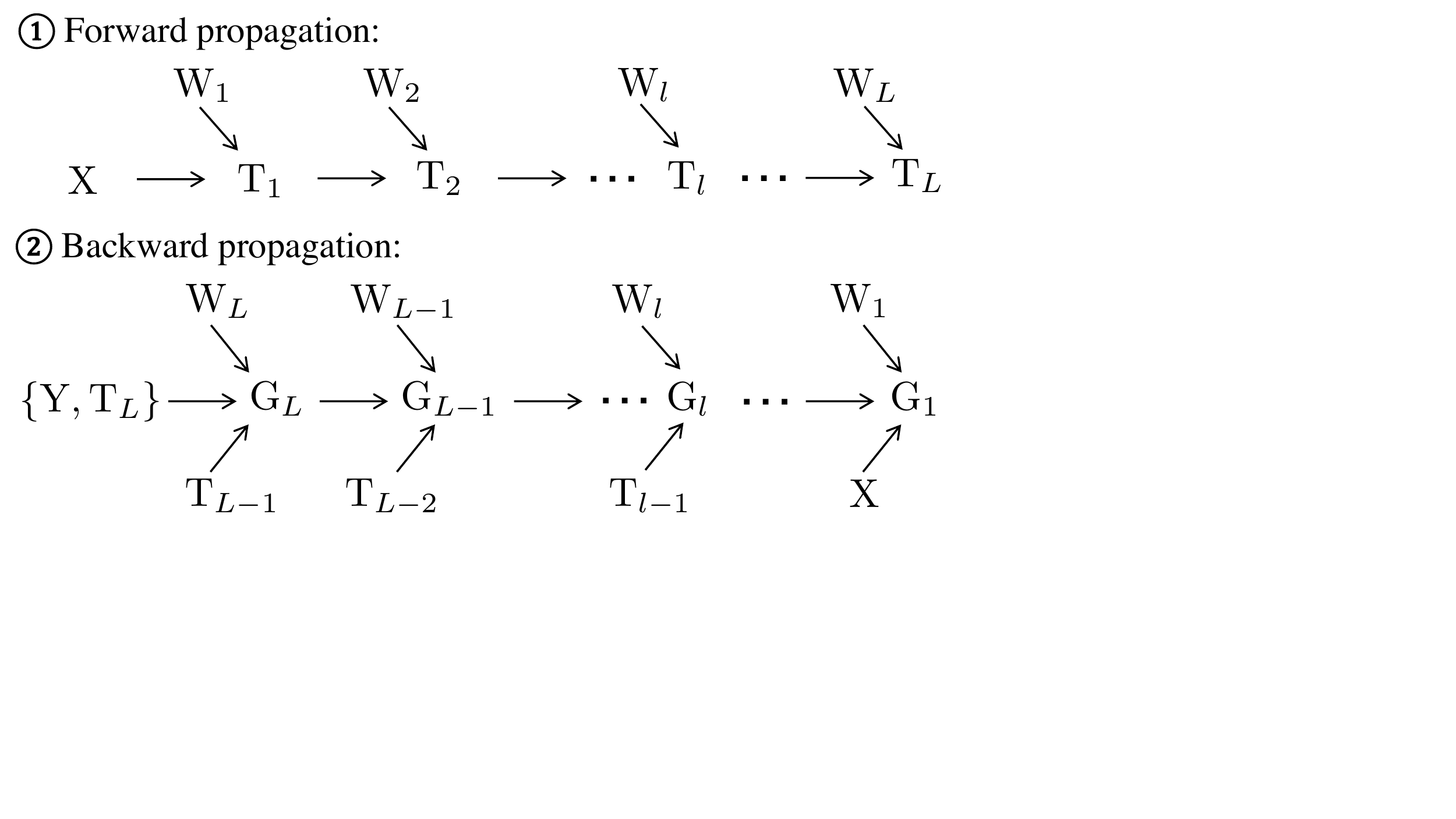}
        \caption{Graphical model of $\raisebox{.5pt}{\textcircled{\raisebox{-.9pt} {1}}}$ forward (from layer $1$ to $L$) and $\raisebox{.5pt}{\textcircled{\raisebox{-.9pt} {2}}}$ backward (from layer $L$ to $1$) propagation: $\mathrm{W}$ shows weights, $\mathrm{G}$ gradients, and $\mathrm{T}$ intermediate representations. }
        \label{fig:markov_chain}
        \vspace{-.4cm}
\end{figure}
To quantify the information flow in forward propagation through DNN layers, previous works analyzed intermediate representations of the input data in a layer-wise manner to understand the flow of information~\cite{montavon2010layer, choromanska2015loss, huang2020layer}.
The visualization of data representations (\ie~layers' outputs) and model's parameters indicate that early layers mostly learn general information about data distribution (\eg~face color), while latter layers learn sample-specific information related to the underlying task (\eg~face identity)~\cite{zeiler2014visualizing, mahendran2015understanding, yosinski2015understanding}.
For theoretical justifications of such analyses, the widely adopted measure has been Shannon's mutual information (MI or $I(\cdot;\cdot)$, hereinafter)~\cite{shannon1948mathematical}. Specifically, for an observed data sample $\mathrm{X}$, labelled with $\mathrm{Y}$, the layer-by-layer computations (from layer $1$ to $L$) form a graphical model (Figure~\ref{fig:markov_chain}-\textcircled{\raisebox{-0.4pt}{1}}). Let $\mathrm{T}$ denote the intermediate representations extracted by a DNN.
According to DPI~\cite{tishby2000information}, $I(\mathrm{X};\mathrm{T})$ and $I(\mathrm{Y};\mathrm{T})$ should \emph{not} increase during the forward propagation, which is extensively studied for DNNs in~\cite{tishby2000information, goldfeld2019estimating, shwartz2017opening}, where it is shown how to quantify information flow in forward propagation. 
In order to apply this estimation, the implicit assumption is that weights $\mathrm{W}$ do not include prior information about one specific $\mathrm{X}$ and the corresponding set of $\mathrm{T}$. 
However, it is not straightforward to apply DPI for analyzing the information flow in backward propagation, because the computation of gradients $\mathrm{G}_l$ in backward propagation form a more complicated graphical model~(Figure~\ref{fig:markov_chain}-\textcircled{\raisebox{-0.5pt}{2}}). Basically, each $\mathrm{G}_{l}$ is computed based on i) the gradients of the next layer $\mathrm{G}_{l+1}$ , ii) the intermediate representations of the former layer $\mathrm{T}_{l-1}$, and iii) the weights of the current layer $\mathrm{W}_{l}$. Thus, the Markov dependencies in backward propagation is much more complicated than forward propagation; especially the fact  that $\mathrm{G}$ is computed based on $\mathrm{W}$ in current iteration and then  $\mathrm{G}$ is used to update $\mathrm{W}$ for the next iteration.

\section{Problem Formulation}

\subsection{Terminology}
We group {\em private information} in two categories. First, ~\textbf{original} information, \ie~the explicit sample $\mathrm{x}\sim\mathrm{X}$, such as a face image of the data owner. Second, \textbf{latent} information, \ie~an implicit attribute $\mathrm{p}\sim\mathrm{P}$ of the data owner, such as gender, age, or race. 
We assume that  $\mathrm{X}$ is a {\em continuous} random variable and $\mathrm{P}$ is a {\em categorical (discrete)} one. The starting point of this distinction between two types of information is inspired by corresponding attacks: \ie~{\em data reconstruction attack}~(DRA) on $\mathrm{X}$, and {\em attribute information attack}~(AIA) on $\mathrm{P}$; as each attack follows different objectives. We define a {\em predictive family} $\mathcal{V}$ as a set of functions that are used for launching an attack (either DRA or AIA).

\subsection{Usable information motivation} \label{sec:use_inf}

Measuring Shannon's MI~\cite{shannon1948mathematical} relies on the implicit assumption that ``attackers are computationally unbounded''. Consequently, in accordance with the DPI, the information in the consequent layers can only decrease. But in practice, the computational power and also the knowledge of attackers might be limited. One can then reasonably expect that the latter layers in a model are able to extract features that are more useful for the target task (\ie~the classification at hand), so that usable information for the task $\mathrm{Y}$, captured in the gradients, should be \emph{increasing} throughout the layers. Similarly, one may also assume that another private attribute $\mathrm{P}$, besides the main task $\mathrm{Y}$, is also easier to extract from the latter layers.

The concept of {\em usable information}~\cite{xu2020theory} was recently introduced as a generalization of Shannon's MI to account for both computational constraints and empirical evidences in ML. Usable information relies on a {\em predictive family} $\mathcal{V}$, the definition of which allows incorporating the computational power of a specific (set of) attackers. More importantly, usable information relaxes the DPI condition and takes into account that in practice we can extract more information from the data by processing it using some ML models compared to observing the raw data. Notice that, this does not imply that less latent information is present in earlier layers, but earlier layers require the attacker to perform heavier computations on the observed data to infer $\mathrm{P}$. 

We explain usable information through an example. Consider a reader of a Google review who attempts to understand the reviewer's sentiment. The DPI states that when the text is processed by a model it cannot have a higher MI with the reviewer's sentiment than the original text. While this holds true for a reader with unbounded computational power, in practice the computational power is limited. Thus, a well-trained sentiment classifier can process the text in such a way that it becomes easier for the reader to extract the sentiment; thereby increasing the usable information through processing. A similar motivation holds for gradients. In accordance with the DPI, observed gradients, in theory, include lots of information about the private data, but practical attacks utilize ML models with limited computational power to extract the desired information. The gradients may contain information in different stages of processing, \eg gradients of later layers may have processed the data in such a way that extracting information from them is easier for an attacker with a low computational budget. In the following, we provide methods for measuring the usable information for AIA and DRA.

\subsection{Latent information leakage: AIA} 
\label{sec:attack_latent}

For analysing AIA, we can almost directly apply the definition of {\em usable information} proposed in~\cite{xu2020theory}; as follows.
\begin{definition}[Usable \textbf{Attribute} Information~\cite{xu2020theory}] Let $\mathrm{G}$ and $\mathrm{P}$ be random variables denoting the {\em observed gradients} and {\em private latent attribute,} and $\mathrm{g}$ and $\mathrm{p}$ be their realisations, respectively. Let $\mathcal{V} =\{f_1,f_2,\dots\}$ denote the AIA's {\em predictive family} of computationally bounded {\em attackers}. For every $f\in\mathcal{V}$, let $f[\mathrm{g}](\mathrm{p})$ denote the probability of $\mathrm{P}=\mathrm{p}$ conditioned on $\mathrm{G}=\mathrm{g}$ 
under model $f$. Similarly, let $f[\emptyset](\mathrm{p})$ denote the prior probability of $\mathrm{P}=\mathrm{p}$ without observing gradients\footnote{This may represent attackers that use some prior or a random guess.}. Let dataset $\mathcal{D} = \{(\mathrm{g}_i,\mathrm{p}_i)\}_{i=1}^N$ denote a set of joint samples of $\mathrm{G}$ and $\mathrm{P}$.

The {\em usable latent information from $\mathrm{G}$ to $\mathrm{P}$ under $\mathcal{D}$} is:
\begin{align}
    \hat{I}_{\mathcal{V}}(\mathrm{G}\rightarrow \mathrm{P}; \mathcal{D}) &= \inf_{f \in \mathcal{V}} \frac{1}{|\mathcal{D}|} \sum_{\mathrm{p}_i \in \mathcal{D}} -\log f[\emptyset](\mathrm{p}_i)\nonumber \\
    & ~~~~~ - \inf_{f \in \mathcal{V}} \frac{1}{|\mathcal{D}|} \sum_{\mathrm{g}_i, \mathrm{p}_i \in \mathcal{D}} -\log f[\mathrm{g}_i](\mathrm{p}_i), 
\label{eq:v_info_AIA}
\end{align}
where $\inf$ is computed as the empirically best-performing attack function in $\mathcal{V}$\footnote{As a technicality, one needs to ensure that there exists an $f'\in\mathcal{V}$ such that both $f'[\mathrm{g}]=Q$ and $f[\emptyset]=Q$ for some probability distribution $Q$, \ie~the model can ignore the side information, so that the definition maintains non-negativity (see the definition of ``optional ignorance'' in~\cite{xu2020theory}).}.
Note that $\inf$ over $\mathcal{V}$ can be empirically achieved by optimization methods like SGD; similar to~\cite{xu2020theory}.

\end{definition}
In the above definition, we remark that each sample $\mathrm{g}_i$ is associated with a  label $\mathrm{p}_i$ which the attacker does not know but tries to infer; to measure the usable information we use the probability that the attacker's model $f$ assigns to observing this label.  As a final remark, membership inference attack, another well-known attack that infers the participation of a particular client in the training dataset~\cite{nasr2019comprehensive, shokri2017membership}, can also be formulated as latent attribute when defining $\mathrm{P}=1$ as membership and $\mathrm{P}=0$ as non-membership.   

\mysubsubsection{Computing $f[\mathrm{g}](\mathrm{p})$ for latent information.}
In our experiments, we consider binary\footnote{We remark that this can be easily generalized to categorical attributes~(\ie~red, yellow, and green) since we can convert the attack objective to be binary: with the attribute or without the attribute~(\ie~red or not red, etc.)} attributes $\mathrm{P} \in \{0,1\}$. AIAs are assumed to have access to an \emph{auxiliary} dataset, including sample points with (\ie $\mathrm{P}=1$) and without (\ie $\mathrm{P}=0$) the target attribute; and thus able to collect a set of gradients $\{(\mathrm{g}_{i},\mathrm{P}=0)\}_{i=1}^{N0}$ and $\{(\mathrm{g}_{i},\mathrm{P}=1)\}_{i=1}^{N1}$. 
Using these two sets of gradients, an AIA attack model $f\in \mathcal{V}$ is trained as a binary classifier with a {\em sigmoid} output layer, which is a solution to
\begin{align}
      \argmin_{f} \  -\mathrm{P} \log \big( f[\mathrm{g}](\mathrm{P})\big).
\label{eq:aia_def}
\end{align}
As we use the cross-entropy loss function, we can interpret the output of $f$ as a probability of $\mathrm{P}$~(see~\cite{xu2020theory} and Section 4.2 of \cite{bishop2006pattern}). 

Thus, after training the binary classifier  $f$ on the auxiliary dataset, we can use $f$ to compute $f[\mathrm{g}_i](\mathrm{p}_i)$ in Equation~\eqref{eq:v_info_AIA}. 
Finally, we set $f[\emptyset](\mathrm{p}_i)=\frac{1}{2}$ for both $\mathrm{P}=1$ and $\mathrm{P}=0$, \ie~we consider a non-informative prior on $\mathrm{P}$.

\subsection{Original information leakage: DRA}
For analysing DRA, we cannot directly use~\eqref{eq:v_info_AIA}, because computing the probability of a sample $x\sim \mathrm{X}$ is computationally challenging due to the high dimensionality of the input variable $\mathrm{X}$. 
Thus, we define the attacker's objective as follows:

\begin{definition}[DRA's Objective]
Suppose we are given a threshold value $\tau \in \mathbb{R}^{+}$ and a sample of the original data $\mathrm{x}\sim \mathrm{X}$. The attacker's reconstruction of the original data is given by the random variable $\mathrm{\hat X}$. Let $\mathrm{\hat x}\sim \mathrm{\hat X}$. We define the random variable $\mathrm{O}^\tau \triangleq \boldsymbol{1}_{\mathsf{sim}(\hat{\mathrm{x}},\mathrm{x})\geq\tau}$ as the {\em DRA's objective}; where $\mathsf{sim}(\cdot,\cdot)$ is a chosen similarity measure, and $\boldsymbol{1}_{()}$ is the indicator function such that $\mathrm{O}^\tau=1$ if the desired similarity threshold $\tau$ is achieved, and $\mathrm{O}^\tau=0$ otherwise.
\end{definition}
Notice that in the above definition, the randomness of $\mathrm{O}^{\tau}$ is over the randomized nature of the reconstruction attack for generating $\hat{\mathrm{X}}$. 

We therefore modify usable information in~\eqref{eq:v_info_AIA}: instead of measuring {\em the probability of the original data given the gradients}, $f[\mathrm{g}_i](\mathrm{x}_i)$, we measure {\em the probability of reconstructing the original data with an objective \ie~to achieve a certain similarity $f[\mathrm{g}_i](\mathrm{O}^{\tau}_i=1)$}.
The modified usable information is then: 
\begin{definition}[Usable \textbf{Original} Information~\cite{xu2020theory}] Let $\mathrm{G}$ be a random variable denoting the {\em observed gradients} and let $\mathrm{X}$ be the random variable denoting the {\em input data sample}. Consider $\mathrm{O}^\tau$, as defined above, a random variable showing the outcome of a DRA. Let dataset $\mathcal{D} = \{\big(\mathrm{x}_i,\mathrm{g}_i\big)\}_{i=1}^N$ denote a set of joint samples of $\mathrm{X}$, $\mathrm{G}$.
Let $\mathcal{V}$ denote the DRA's {\em predictive family}. 
The {\em usable information from $\mathrm{G}$ to $\mathrm{X}$ under $\mathcal{D}$}
\begin{align}
    \hat{I}_{\mathcal{V}}(\mathrm{G}\rightarrow \mathrm{X}; \mathcal{D}) &= \inf_{f \in \mathcal{V}}\frac{1}{|\mathcal{D}|} \sum_{\mathrm{x}_i\in \mathcal{D}} -\log f[\emptyset](\mathrm{x}_i) \nonumber \\
    &~~~~~ - \inf_{f \in \mathcal{V}} \frac{1}{|\mathcal{D}|} \sum_{(\mathrm{g}_i,\mathrm{x}_i)\in \mathcal{D}} -\log f[\mathrm{g}_i](\mathrm{x}_i),
\label{eq:v_info_DRA_1}
\end{align}
is approximated by
\begin{equation}
\begin{split}
 \hat{I}_{\mathcal{V}}(\mathrm{G}\rightarrow \mathrm{X}; \mathcal{D}) \leq \inf_{f \in \mathcal{V}} \frac{1}{|\mathcal{D}|} \sum_{\mathrm{x}_i\in \mathcal{D}} -\log f[\emptyset](\mathrm{O}^{\tau}_i=1) \\
 -\inf_{f \in \mathcal{V}} \frac{1}{|\mathcal{D}|} \sum_{(\mathrm{g}_i,\mathrm{x}_i)\in \mathcal{D}} -\log f[\mathrm{g}_i](\mathrm{O}^{\tau}_i=1),
\end{split}
\label{eq:v_info_DRA_2}
\end{equation}
where we remark that $\mathrm{O}_i$ is a function of $\mathrm{x}_i$ as it depends on the original data and hence the summation being over $\mathrm{x}_i$. 
\end{definition}

To this end, we need a method to compute $f[\mathrm{g}](\mathrm{O}^\tau=1)$, the probability that attack model $f$ in $\mathcal{V}$ can successfully reach the objective $\mathsf{sim}(\hat{\mathrm{x}}_i ,\mathrm{x}_i) \geq \tau$, given the gradients $\mathrm{g}_i$; compared to $f[\emptyset](\mathrm{O}^\tau=1)$, the best random guess (in practice a random reconstruction as we will discuss next). Thus, computing \eqref{eq:v_info_DRA_2} serves as a proxy to approximate \eqref{eq:v_info_DRA_1}, which in practice is difficult, if not impossible, to compute. We remark that \eqref{eq:v_info_DRA_2} is an approximation based on the chosen similarity measure and prescribed thresholds. This approximation is neither a lower-bound nor an upper-bound for \eqref{eq:v_info_DRA_1}. The error of such approximation relies on how close one similarity measure is with the real density measure in $f[\cdot](\mathrm{x}_i)$. 

\mysubsubsection{DRA in practice}. To reconstruct the original data, state-of-the-art DRAs~\cite{zhu2019deep, zhao2020idlg, geiping2020inverting} usually start by randomly initializing dummy data and feeding it into the model to get dummy gradients. Then, the dummy data is optimized such that the dummy gradients get close to the real gradients.
More specifically, the DRA tries to minimize the following objective by updating both dummy input $\mathrm{x}'$ and dummy label $\mathrm{y}'$,
\begin{align}\label{eq:DRA_objective}
    \hat{\mathrm{x}}, \hat{\mathrm{y}} = 
    \argmin_{\mathrm{x}', \mathrm{y}'} \ \mathrm{dist}( \hat{\mathrm{g}},\mathrm{g}),
\end{align}
where $\hat{\mathrm{g}}$ and $\mathrm{g}$ denote the dummy and observed gradients, respectively, and $\mathrm{dist}(\cdot, \cdot)$ can be any suitably chosen distance metric for them. For example, \cite{zhu2019deep} and \cite{zhao2020idlg} use $\|\hat{\mathrm{g}} - \mathrm{g}\|_p$ with $p$-norm, while \cite{geiping2020inverting} uses cosine similarity between $\hat{\mathrm{g}}$ and $\mathrm{g}$.
The attack will be successful if minimizing the corresponding distance will result in the dummy input data being close to the real private information, which can reach high similarity (\eg~to a pixel-wise level for images) as shown in~\cite{zhu2019deep,zhao2020idlg,geiping2020inverting}. Note that $\hat{\mathrm{x}}$ and $\hat{\mathrm{y}}$ can be one individual data point or a batch of data points. For a batch size $B>1$, the optimizer can have $B!$ different permutations, so that choosing the right gradient descent direction is complicated - as a solution a single training sample can be updated in each optimization step~~\cite{zhu2019deep,zhu2020r}. It may fail when gradients have sufficient aggregation (\eg~a large batch size) because the optimization becomes hard to solve (\ie~more variables to optimize over than we have constraints).

\mysubsubsection{Computing $f[\mathrm{g}](\mathrm{O}^\tau=1)$ for original information.}\label{subsec:original}
In order to approximate $\hat{I}_{\mathcal{V}}(\mathrm{G}\rightarrow \mathrm{X}; \mathcal{D})$, we have to
i) define what similarity measure to use; \eg~structural similarity index measure (SSIM),
ii) define the threshold for the DRA's objective; \eg~SSIM$\geq \tau=0.5$, and
iii) approximate $f[\mathrm{g}](\mathrm{O}^{\tau}=1)$; \eg~$f[\mathrm{g}](SSIM\geq 0.5)\approx 0.2$. 
The similarity metric can be any suitable metric that measures the similarity between original and reconstructed data, depending on the data types (\eg~images or audios) and the focus (\eg~better realization for machines or humans). As one example, both peak signal-to-noise ratio (PSNR) and SSIM can be used for images; where the latter is defined based on the human visual system and thus is more suitable for measuring human perception~\cite{wang2004image, hore2010image}, while the former is a general statistical measure in signal processing applications. 
Here we consider using SSIM given by $\mathtt{SSIM}(\hat{\mathrm{x}}, \mathrm{x}) \triangleq l(\hat{\mathrm{x}}, \mathrm{x}) c(\hat{\mathrm{x}}, \mathrm{x}) s(\hat{\mathrm{x}}, \mathrm{x})$, a combination of luminance ($l$), contrast ($c$), and structure ($s$).
$\mathtt{SSIM}$ lies within a range of $[0,1]$, with the value of 1 being achieved for a close-to-perfect reconstruction.
We then approximate $f[\mathrm{g}](\mathrm{O}^\tau=1)$ as follows: i) for one observed gradient $\mathrm{g}_j$, we randomly initialize attack models in \eqref{eq:DRA_objective} for $R$ rounds; ii) for each round we perform the optimization as discussed in the attack model and compute $\mathtt{SSIM}$ to obtain $(\mathrm{g}_j,\mathtt{SSIM}_i)$, iii) using these $R$ SSIM values, $\{(\mathrm{g}_j,\mathtt{SSIM}_i)\}_{i=1}^R$, we compute the probability of obtaining $\mathtt{SSIM}_i\geq \tau$ as:
\begin{align}
    f[\mathrm{g}](\mathrm{O}^\tau=1) = \frac{\sum_{i=1}^R\boldsymbol{1}_{(\mathtt{SSIM}_i\geq \tau )}}{R}.
\end{align}
When having no side information, the best one could do is to reconstruct a random image; similarly to the above, one can compute in how many rounds $\mathtt{SSIM}_i\geq \tau$, $i=1,...,R$. 
To use this probability estimate in the usable original information, we repeat the attack $R$ times for each sample $\mathrm{x}$ in $\mathcal{D}$.
While computationally expensive, such a framework allows us to compute a probabilistic estimate of a successful reconstruction attack.

\section{Sensitivity Measure on Private Information} \label{sec:sensi_meas}

Having defined our {\em usable information} measure for quantifying the amount of private information in the gradients in the previous section, 
we now explain how to reason about the leakage based on the {\em gradient sensitivity} measure and validate the efficacy of usable information.
The sensitivity measure is motivated in two ways: i) it gives further insight into model characteristics that facilitate private information leakages, ii) it allows us to measure information leakage in a way that does not depend on the attack models; and thus, allows us to validate our proposed method without making any explicit assumption on the attackers' objectives. In the following, we explain how  {\em gradient sensitivity}, which previously has been studied for examining model robustness and generalization~\cite{novak2018sensitivity, ding2018sensitivity}, can be expanded into metrics to be used for studying both original and latent information.

\subsection{Sensitivity justification}
Consider again the objective in Equation~\eqref{eq:DRA_objective}. In cases where the objective function $\text{dist}( \hat{\mathrm{g}},\mathrm{g})$ has multiple or very flat minima, solving the optimization; and thus, recovering the true original data reconstruction, will be more challenging as multiple solutions with similar objective function values may exist. 
Therefore, the success of data reconstruction depends on the structure of the objective function. Previous works have introduced rank-based metrics to quantify the ability to find the unique optimizer~\cite{zhu2019deep} relying on the fact that the optimization problem can be rewritten as a system of equations. If more parameters exist than equations, no unique solution will exist. 
Alternatively, one could choose a metric that accounts for the flatness of the loss function (and hence the gradients) in the parameter or input space. As these flat minima are a common occurrence in the loss functions of DNNs~\cite{draxler2018essentially}, having a metric that is able to account for this is essential. This notion of sensitivity based on flatness has also been applied in measuring model robustness on adversarial example attacks~\cite{zahavy2016ensemble, ding2018sensitivity}.
A model that is insensitive to certain changes in inputs is expected to have high robustness. On the other hand, by bounding the sensitivity of outputs \wrt~inputs leveraging DP, one can certify model robustness to changes in input~\ie~adversarial examples, proposed in~\cite{lecuyer2019certified}.

\subsection{Sensitivity of gradients \wrt~original information}

For quantifying the original information leakage, the {\em gradients sensitivity} metric is quite straightforward: we compute the sensitivity of the \emph{gradient} \wrt~the private \emph{input} via the Jacobian matrix of the gradients. We use the following intuition to measure private information leakage: if the gradient sensitivity is low; that is, gradients change insignificantly when the input is altered, reconstructing the correct input will be more challenging as more possible reconstructions within the same similarity level will exist. We will utilize the Jacobian matrix of the gradients \wrt~the input (similar to input-output Jacobian~\cite{novak2018sensitivity, sokolic2017robust}) to reflect how sensitive the gradients are when changing the input, \ie~a sensitivity measure on gradients.
The input-gradient Jacobian is calculated by:
\begin{equation}
        \mathbf{J}_l^{(\bm{G})}(\bm{X}) = \frac{\partial \mathbf{g}_l(\bm{X})}{\partial \bm{X}} = \frac{\partial}{\partial \bm{X}} \left( \frac{\partial \ell(\bm{X},\bm{y}, \bm{W})}{\partial \bm{W}_l} \right),
\end{equation}
where $\mathbf{g}_l(\cdot)$ represents the function that produces layer $l$'s gradient $\bm{G}_l$. 
Besides, $\ell(\cdot)$ is the loss function over $\bm{X}$, ground truth $\bm{y}$, and parameters of the complete model $\bm{W}$, so $\mathbf{g}_l(\cdot)$ can be regarded as the partial derivative of $\ell(\cdot)$ \wrt~layer $l$'s parameters $\bm{W}_l$ (\ie~backward propagation).

Then we compute the Frobenius norm (referred to $F$-norm) of the above input-gradient Jacobian matrix~\cite{novak2018sensitivity}, which indicates the general original information risk. As in our case Jacobians are compared across layers with different sizes, we include two other norms, \ie~$1$-norm and the $\infty$-norm. Using different norms will help in capturing attackers with different capabilities, similarly to~\cite{cisse2017parseval, lecuyer2019certified}, where $p$-norm reflects how the attacker measures distance between two data samples (\eg~$1$-norm reflects all dimensions of the data sample, and $\infty$-norm reflects on one dimension). Thus, given $K$ data samples, we compute the Jacobian with $p$-norm (referred to as `Jacobian $p$-norm' hereinafter) averaged over the data samples as the leakage risk of the original information ($\bm{X}$) in layer $l$'s gradients:
\begin{align}
      \mathcal{R}_l^{(\bm{X})} &= \mathbb{E}_{\Delta\bm{X}}\Big[\| \mathbf{g}_l(\bm{X})-\mathbf{g}_l(\bm{X}+\Delta\bm{X}) \|_p\Big], \nonumber \\
      &= \frac{1}{K} \sum_{k=1}^K \left\lVert \mathbf{J}^{(\bm{G})}_l(\bm{X}_k)\right\rVert_p,
\label{eq:jaco_x}
\end{align}
where $p = F, 1,\text{or}\ \infty$. Note that the size of $\bm{G}_l$ may still have an impact on the computed sensitivity. A larger $\bm{G}_l$ size can result in larger sensitivity in terms of $1$-norm and $F$-norm.
When reconstructing high-dimensional input data, a large $\bm{G}_l$ size is necessary, which can be reflected by these norms.

\subsection{Sensitivity of gradients \wrt~latent information}

For attacks on latent information~\cite{melis2019exploiting}, the success of the attack lies in the ability of the classifier to distinguish between whether certain gradients were computed over data with or without a certain attribute. Specifically, it relies on how accurate the classifier trained on samples with or without an attribute can become (see Section~\ref{sec:attack_latent}). 

Computing the sensitivity of gradients \wrt~latent information is not as easy as \wrt~the original information for two reasons. First, the latent information (the private label) is not explicitly fed into the DNN, and thus the backpropagation computational graph cannot be used to compute $\frac{\partial \bm{\mathrm{g}}}{\partial \bm{p}}$. Second, in Equation~\eqref{eq:jaco_x}, the sensitivity is computed separately over each input sample, but a large number of inputs have the same latent information and the attacker's classifier only produces a binary prediction.  
Thus, the fine-grained sensitivity measure in~\eqref{eq:jaco_x} cannot be directly used to grasp the amount of latent (\ie~coarse) data in the gradients. 

To quantify the changes in gradients, we offer a more coarse-grained approach. First, we separate the target dataset into two parts based on the presence of one specific latent information, and then we compare the subspace distance between gradients computed on these two parts of the dataset. This still follows the same intuition as understanding how sensitive the gradients are \wrt~latent information, but considers a more `high-level' measurement.

Let us assume that a dataset $\mathcal{S}$ consists of two disjoint subsets: $\mathcal{S}_{0}$, whose samples do not have the target latent attribute, and $\mathcal{S}_{\text{1}}$, whose samples have it. Then, we obtain the corresponding gradients of layer $l$ computed on these two subsets by:
\begin{equation}
    \bm{G}_l^{(\text{0})} = \mathbb{E}_{\bm{X} \in \mathcal{S}_{\text{0}}}[\mathbf{g}_l(\bm{X})] \ \text{and} \ \bm{G}_l^{(\text{1})} = \mathbb{E}_{\bm{X} \in \mathcal{S}_{\text{1}}}[\mathbf{g}_l(\bm{X})].
\end{equation}

For each pair of matrices $\bm{G}_l^{(\text{0})}$ and $\bm{G}_l^{(\text{1})}$ (in $l \in \{1,...,L\}$), let $\mathbb{G}_l^{(\text{0})}$ and $\mathbb{G}_l^{(\text{1})}$ denote their corresponding linear subspaces, respectively. To measure the difference between the computed gradients with and without target latent attribute, we can compute the Grassmann geodesic distance~\cite{drmac2000principal, ye2016schubert} between these two subspaces as the leakage risk of this latent information $\bm{p}$ in layer $l$'s gradients:
\begin{equation}
\begin{split}
    \mathcal{R}_l^{(\bm{p})} = \text{dist}(\bm{G}_l^{(\text{0})}, \bm{G}_l^{(\text{1})}) = d_{\mathrm{Gr}(k,n)} (\mathbb{G}_l^{(\text{0})},\mathbb{G}_l^{(\text{1})}) \\
    = \left(\sum\nolimits_{i=1}^k \theta^2_i\right)^{1/2},
\end{split}
\label{eq:jaco_y}
\end{equation}
where $\mathrm{Gr}(k,n)$ denotes the Grassmann manifold, and both $\mathbb{G}_l^{(\text{0})}$ and $\mathbb{G}_l^{(\text{1})}$ are elements of $\mathrm{Gr}(k,n)$ and are $k$-dimensional linear subspace in $\mathbb{R}^n$. $k$ is the layer $l$'s gradient size $N_l \times N_{l-1}$. $\theta_i \ \text{for}\ i \in \{1,...,k\}$ are principal angles between the two subspaces which can be computed using numerical methods~\cite{bjorck1973numerical}.
In this paper we use the Grassmann distance; other common distances defined on Grassmannians (\eg~Asimov)~\cite{ye2016schubert} can be derived similarly.

\section{Numerical Evaluation}
In this section, to validate the proposed {\em usable information}~(Section~\ref{sec:use_inf}) measure, we first provide a comprehensive experimental evaluation using benchmark datasets and DNN architectures, and then we utilize the {\em sensitivity measure}~(Section~\ref{sec:sensi_meas}) to further verify the integrity of our proposed measure.

\subsection{Evaluation setup}

\label{section:model}

\begin{table}[]
\caption{Models and datasets used in our experiments.}
\label{tab:models}
\resizebox{\columnwidth}{!}{%
\begin{tabular}{@{}lll@{}}
\toprule
Model  & Architecture & 
Datasets
\\ \midrule
LeNet~\cite{lecun1998gradient}   & C6(5)-P(2)-C16(5)-P(2)-F120-F84-O                              & \multirow{3}{*}{\begin{tabular}[c]{@{}l@{}}CIFAR-100, \\ LFW, CelebA, \\ \& PubFig\end{tabular}}     \\
AlexNet~\cite{krizhevsky2012imagenet} & C8(5)-C16(3)-P(2)-C32(3)$\times$3-P(2)-F120-F84-O         \\
VGG9~\cite{simonyan2014very}    & C8(3)-C16(3)$\times2$-P(2)-C32(3)$\times3$-P(2)-F256-F128-O \\\midrule
TextClf~\cite{kim2014conv} & C16(3)-P(2)-C16(8)-P(2)-F50-O                              & IMDB \& CSI    \\ \bottomrule
\end{tabular}%
}
\begin{flushleft}
{\scriptsize
Notations. C$i$($j$): a convolutional layer with $i$ 2D filters each of size $j\times j$. F$i$: a fully connected layer with $i$ neurons. P($j$): a max-pooling layer with a window size of $j\times j$. O: the output classification layer with $Y$ classes (depending on the dataset).} \hfill \\
\end{flushleft}
\vspace{-10pt}
\end{table}

\mysubsubsection{Model.}
We perform experiments on four DNN models used by previous works on privacy threats of collaborative learning~\cite{zhu2019deep, zhao2020idlg, geiping2020inverting, melis2019exploiting, nasr2019comprehensive}. See Table\ref{tab:models} for the details of each DNN's architecture.
The number of neurons in the output layer is adjusted based on the number of classes of the dataset used to train.
All models use ReLU activation functions for Conv and FC layers (except the output layer with Softmax).

\mysubsubsection{Datasets.}
We conduct evaluations on six datasets.
LeNet, AlexNet, and VGG9 models are trained on CIFAR-100~\cite{krizhevsky2009learning} and \emph{three other image datasets} with attributes, including Labeled Faces in the Wild (LFW)~\cite{huang2008labeled}, Large-scale CelebFaces Attributes (CelebA)~\cite{liu2015faceattributes}, and Public Figures Face Database (PubFig)~\cite{kumar2009attribute}. 
TextClf is trained on \emph{two text datasets}: IMDB reviews~\cite{maas2011learning} and CSI corpus~\cite{verhoeven2014clips}.
LFW contains 13233 face images (cropped as $62\times47$ RGB). All images are labeled with around 100 attributes such as gender, race, age, hair color, etc. 
We use a subset of the cropped version (\ie~15000 images of $64\times64$ RGB) of CelebA which contains face images of celebrities with 40 attribute annotations such as gender, hair color, eyeglasses, etc.
We also use a cropped version ($100\times100$ RGB) of PubFig which contains 8300 facial images made up of 100 images for each of 83 persons~\cite{pinto2011scaling}, marked with 73 attributes (\eg~gender, race, etc).
In IMDB reviews~\cite{maas2011learning} dataset, each review is labeled with sentiment and length, and in CSI corpus~\cite{verhoeven2014clips}, each corpus review is labeled with sentiment, veracity, length, etc. 

\mysubsubsection{Training setup.}
We conduct our experiments on a cluster with 50 nodes where each has 4 Intel(R) Xeon(R) E5-2620 CPUs (2.00GHz), one/two NVIDIA RTX 6000 GPU(s) (24GB), and 24GB/48GB DDR4 RAM. Pytorch v1.4.0 is used for both usable information (\eg~conducting DRAs and AIAs) and sensitivity computations.
We refer to the codes for more details of gradient change measures and~\cite{zhao2020idlg, melis2019exploiting} for attack settings.

\subsection{Layer-wise localization of information leakages}

\mysubsubsection{Predictive Families.} 
We define a predictive family that covers the most recent attacks including~\cite{zhu2019deep, zhao2020idlg, geiping2020inverting, melis2019exploiting} to compute the empirical usable information as in Equation~\eqref{eq:v_info_AIA} and~\eqref{eq:v_info_DRA_2}.
In particular, for original information, the predictive family includes deep learning techniques with the same training hyperparameters used in the previous attacks~\cite{zhu2019deep, zhao2020idlg, geiping2020inverting}. Moreover, for the latent information, random forests are included in the predictive family with the hyperparameters tuned in a state-of-the-art manner based on previous successful attacks~\cite{melis2019exploiting}.  In general, to empirically achieve the Infima over the predictive family in Equation~\eqref{eq:v_info_AIA} and~\eqref{eq:v_info_DRA_2}, one needs not only to include the state-of-the-art attacks in the literature but also to perform hyperparameter tuning for learning better attack models. We remark that any new-coming attacks can be added to the predictive family and their results can be merged into current results without redoing the computation of previous attack models.

\begin{figure*}[t!]
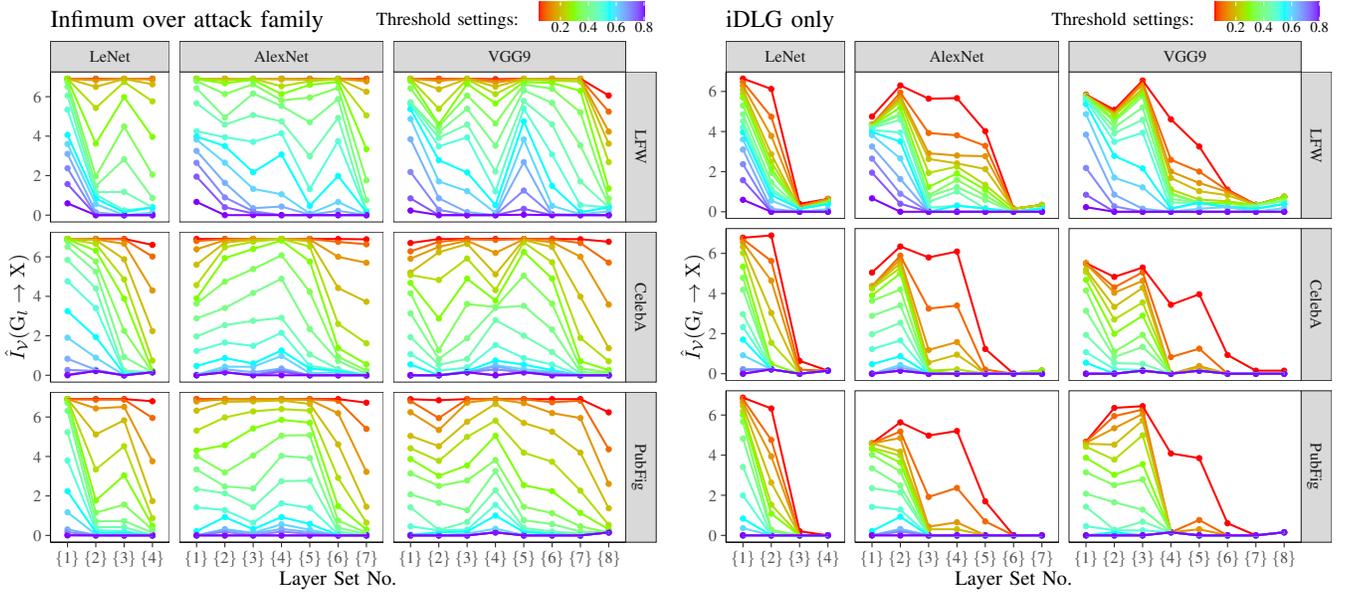

    \centering
    \begin{subfigure}{1\columnwidth}
    \iftex
        \resizebox{!}{0.9\columnwidth}{\input{figs/thres_merged.tex}}
    \else
        \includegraphics[width=1\columnwidth]{figs/thres_merged.pdf}
    \fi
    \end{subfigure}
    \begin{subfigure}{1\columnwidth}
    \iftex
        \resizebox{!}{0.9\columnwidth}{\input{figs/thres_dlg.tex}}
    \else
        \includegraphics[width=1\columnwidth]{figs/thres_dlg.pdf}
    \fi
    \end{subfigure}
    \caption{Original information leakage measured by usable information with thresholds from 0.05 to 0.8 (with a step size of 0.05) in each 2-layer set's. Left figure shows the results of Infimum over predictive family~\cite{zhu2019deep, zhao2020idlg, geiping2020inverting}. Right figure shows the results of iDLG model only~\cite{zhao2020idlg}.}
    \label{fig:thres_vinfox}
    \vspace{-.3cm}
\end{figure*}

\mysubsubsection{Usable original information.}
To measure layer-wise information leakage, we perform attacks on individual layers. First, we found that all existing DRAs fail in achieving any meaningful outcome when using gradients of only one layer. We thus measure the leakage on the set of at least two consecutive layers at a time. Specifically, in Figure~\ref{fig:thres_vinfox}, set $\{1\}$
denotes layers $1$ and $2$, set $\{2\}$ denotes layers $2$ and $3$, \emph{etc}. 
Figure~\ref{fig:thres_vinfox}~(Left) shows the results of the results of the predictive family (measured with $\inf$), which includes DLG~\cite{zhu2019deep}, iDLG~\cite{zhao2020idlg}, and Inverting Gradients~\cite{geiping2020inverting}. In Figure~\ref{fig:thres_vinfox}~(right), we show the results of a single attack: iDLG attack model~\cite{zhao2020idlg}.

The results of the predictive family, including three attacks, indicate that for LeNet the leakage is monotonically decreasing for thresholds larger than 0.5, but for deeper DNNs, \ie~AlexNet and VGG9, the leakage is highest in the middle layer sets.
The iDLG attack~\cite{zhao2020idlg} model shows a little different pattern that the original information leakage risk generally decreases when moving from the first 2-layer set to the last 2-layer set. 
This could probably be due to that the cosine similarity used in Inverting Gradients~\cite{geiping2020inverting} for the cost function could better capture gradient differences than the euclidean distance used by~\cite{zhao2020idlg}, especially for middle layers in more complicated DNN architectures. Thus, Inverting Gradients outperforms iDLG in most cases, so that the former will represent the final original information leakage over the defined predictive family.
Still, one common observation is that for both attack models the last layer-set has the lowest level of information leakage. 
\begin{figure*}[t!]
    \centering
    \begin{subfigure}{1.6\columnwidth}
    \iftex
        \resizebox{1\columnwidth}{!}{\input{figs/vinfo_y.tex}}
    \else
        \includegraphics[width=1\columnwidth]{figs/vinfo_y.pdf}
    \fi
    \end{subfigure}
    \begin{subfigure}{0.4\columnwidth}
        \includegraphics[width=1\columnwidth]{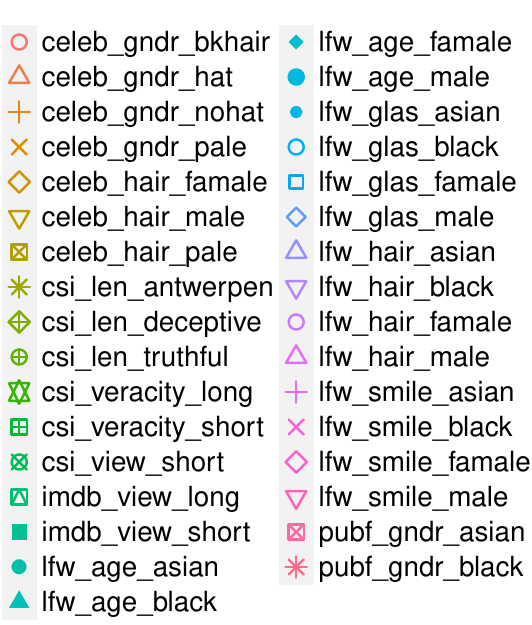}
        \vspace{5pt}
    \end{subfigure}
    \vspace{-5pt}
    \caption{Latent information leakage, measured by usable information, in each layer's gradients. Scattered points refer to private attributes being measured for different datasets and classification tasks. Blue lines give the \texttt{LOESS} regression curve. Dashed lines (\greydashedline) separate the feature extractor and the classifier of the model.}
    \label{fig:vinfo_y}
    \vspace{-.3cm}
\end{figure*}

\mysubsubsection{Usable latent information.}
For latent information, we additionally train the TextClf on the two text datasets and put CIFAR100 aside because it does not have multiple attributes. 
Also, since each dataset can have multiple types of attributes considered as private latent information, we compute usable information on all these (private) attributes and then fit smooth curves using nonparametric local weighted (\ie~\texttt{LOESS}) regression
to clarify the general trend across layers. 
We also normalize the computed usable information of one model with the maximum value of its layer for comparability across models and tasks. 
As in Figure~\ref{fig:vinfo_y}, the results show that the latent information leakage risk follows a trend that increases when moving through the feature extractor layers, reaches its maximum at the first classifier layer, and then decreases. This tendency is similar across all models. 
In addition, the results based on usable information also confirm our analysis in Section~\ref{sec:MI} that the private information flow in gradients does not satisfy DPI.

\subsection{Sensitivity analysis}
We continue with a sensitivity analysis. 

\begin{figure*}
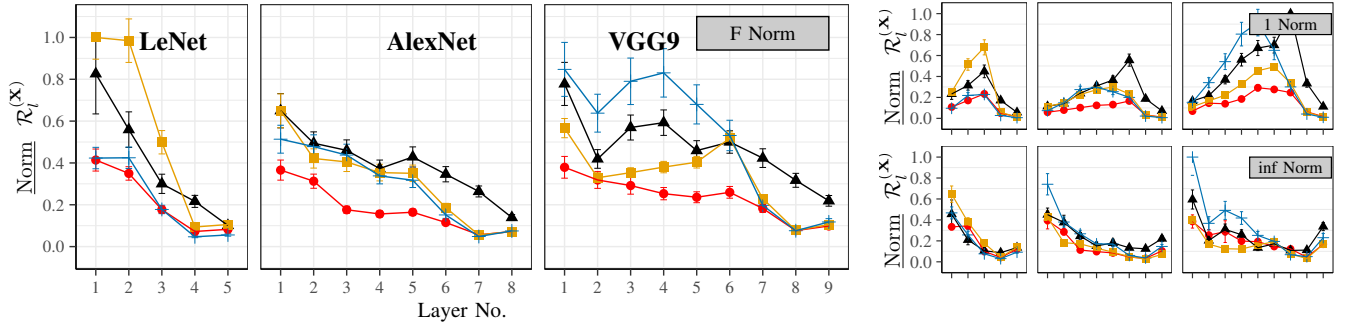

    \begin{minipage}[c][][c]{1.3\columnwidth}
        \iftex
            \resizebox{1\columnwidth}{!}{\input{figs/changes_sen_p2.tex}}
        \else
            \includegraphics[width=1\columnwidth]{figs/changes_sen_p2.pdf}
        \fi
    \end{minipage}
    \begin{minipage}[c][][c]{0.7\columnwidth}
        \iftex
            \resizebox{1\columnwidth}{!}{\input{figs/changes_sen_p1.tex}}
        \else
            \includegraphics[width=1\columnwidth]{figs/changes_sen_p1.pdf}
        \fi
        \iftex
            \resizebox{1\columnwidth}{!}{\input{figs/changes_sen_pi.tex}}
        \else
            \includegraphics[width=1\columnwidth]{figs/changes_sen_pi.pdf}
        \fi
        \vspace{5pt}
    \end{minipage}
    \vspace{-5pt}
    \caption{Original information leakage risks, measured by sensitivity in each layer's gradients, based on CIFAR100 (\blackline), LFW (\blueline), CelebA (\redline), PubFig (\orangeline) dataset. Error bars are 95\% confidence intervals.}
    \label{fig:changes_sen}
\end{figure*}

\mysubsubsection{Sensitivity of gradients \wrt~original information.}
We compute Equation~\eqref{eq:jaco_x} to measure original information leakage risks. 
As shown in Figure~\ref{fig:changes_sen} (Left) for the $F$-norm, the sensitivity in overall shows a similar pattern as the usable information with the attack model of~\cite{zhao2020idlg} in Figure~\ref{fig:thres_vinfox}. 
Specifically, gradients of the first layers are more sensitive to changes in inputs compared to the latter layers, thus they could potentially carry more private original information that can be detected by attackers easier. 
Besides, $1$-norm sensitivity results, in Figure~\ref{fig:changes_sen} (Right-Top), have a similar pattern with the attack model of~\cite{geiping2020inverting}, where DNNs' middle parts, \ie~around the connection from feature extractor to classifier, get the highest information leakage. 
We know that the difference for attacks on usable information probably is due to the distance measure in their cost functions, \ie~iDLG~\cite{zhao2020idlg} used $2$-norm while Gradient Inverting attack model~\cite{geiping2020inverting} used cosine similarity. This information leakage may be tied with the sensitivity of gradients \wrt~the input (\ie~private information) with $p$-norm where $p$ links to attack models with different capabilities.

\begin{figure*}
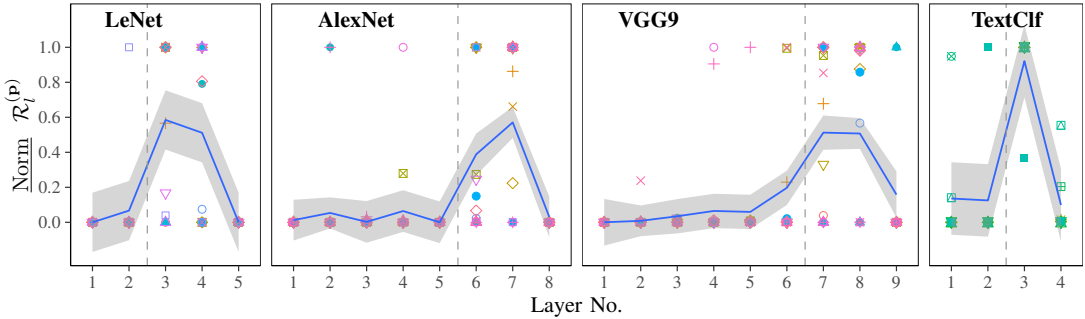

    \centering
    \iftex
        \resizebox{1.65\columnwidth}{!}{\input{figs/changes_sim.tex}}
    \else
        \includegraphics[width=1.65\columnwidth]{figs/changes_sim.pdf}
    \fi
    \vspace{-5pt}
    \caption{Latent information leakage risks, measured by subspace distances, in each layer's gradients. Scattered points correspond to private attributes. Blue lines give the \texttt{LOESS} regression. Dashed lines (\greydashedline) refer to the feature extractor-classifier connection.
    }
    \label{fig:changes_sim}
    \vspace{-.3cm}
\end{figure*}

\mysubsubsection{Sensitivity of gradients \wrt~latent information.}
We compute Equation~\eqref{eq:jaco_y} to measure latent information leakage risks. 
Similar to 
usable information, we measure all possible attributes of used datasets and again plot the regression curves for each model.
As Figure~\ref{fig:changes_sim} shows, $\mathcal{R}^{(\bm{p})}_l$ generally follows a similar trend as the latent information risks computed using usable information. That is, the layers related to the classifier have higher leakage risks than the feature extractor layers. Most risks are either in the classifier's first layer or the second layer.
While this is slightly different from the results in Figure~\ref{fig:vinfo_y} where the first layer always has the highest leakage, the overall trend remains the same. 
We also notice that several scattered points closely coincide, and the last layer has a lower risk compared to that in usable information. One explanation is that this measure may tend to capture the extreme risks among layers so the first and second layers of the classifier show significantly higher leakage risks than the other layers.
Overall, the computation measuring gradient changes (\ie~sensitivity) depends only on the gradients, is easy-to-compute and captures the main pattern of layer-wise leakage risks measured by attack-based usable information.

\subsection{Impacts of training hyperparameters}

We now analyze how some training hyperparameters in collaborative learning affect gradient-based information leakages. For computing original information leakage, we fix the threshold to evaluate the expected performance of attack models for different hyperparameter values.
In particular, we fix the threshold to the expectation of attack outcomes' probability distribution; then for one individual attack on one sample we have $o_T=E_{\mathrm{o} \sim \mathcal{A}}(f[\mathrm{g}])$. In such a way, we evaluate the expected performance of one attack on $\mathrm{g}$. 

\begin{figure*}[t]
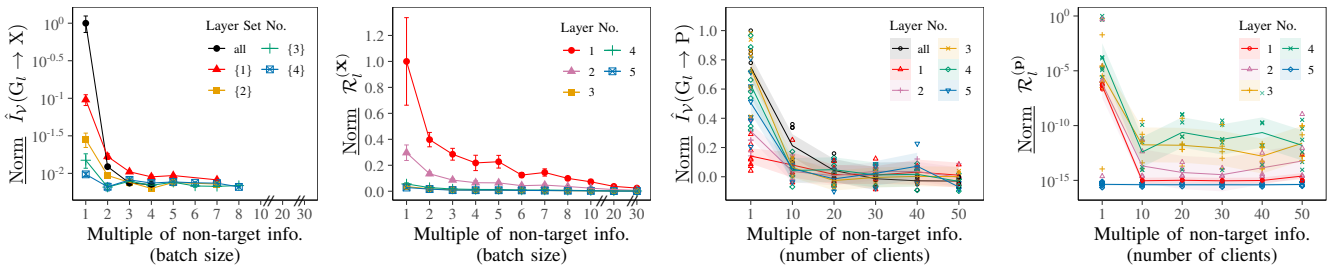

    \centering
    \begin{subfigure}{0.5\columnwidth}
        \iftex
          \resizebox{1\columnwidth}{!}{\input{figs/aggr_vinfo_x}}
        \else
            \includegraphics[width=1\columnwidth]{figs/aggr_vinfo_x.pdf}
        \fi
    \end{subfigure}%
    \begin{subfigure}{0.5\columnwidth}
        \iftex
            \resizebox{1\columnwidth}{!}{\input{figs/aggr_changes_sen}}
        \else
            \includegraphics[width=1\columnwidth]{figs/aggr_changes_sen.pdf}
        \fi
    \end{subfigure}%
    \begin{subfigure}{0.5\columnwidth}
        \iftex
            \resizebox{1\columnwidth}{!}{\input{figs/aggr_vinfo_y}}
        \else
            \includegraphics[width=1\columnwidth]{figs/aggr_vinfo_y.pdf}
        \fi
    \end{subfigure}
    \begin{subfigure}{0.5\columnwidth}
        \iftex
            \resizebox{1\columnwidth}{!}{\input{figs/aggr_changes_sim}}
        \else
            \includegraphics[width=1\columnwidth]{figs/aggr_changes_sim.pdf}
        \fi
    \end{subfigure}
    \caption{Influence of gradient aggregation on risks of original information [Left two] and latent information [Right two], measured on LeNet trained on CIFAR100 and LFW, respectively (Note: Points corresponding to failed trials of attacks are not plotted. Logarithmic scale is used in some plots for better visualization).}
    \label{fig:aggr}
\end{figure*}

\mysubsubsection{Aggregation levels.}
Performing aggregation before sharing gradients with others is a common strategy in collaborative learning. This could be the \emph{local} aggregation (\eg~batch size in \texttt{FedSGD} or multiple steps in \texttt{FedAvg}) and also the \emph{global} aggregation over updates of all clients. 
For our aggregation measurement of one target's (victim) original or latent information, we mix it with multiple gradients of non-target information, \ie~information irrelevant to the private information targeted by attackers.
Our empirical results show that aggregation can significantly reduce gradients' information leakage risks (see Figure~\ref{fig:aggr}). 
Specifically, disclosing target original information from gradients that have been updated on non-target information with an amount 10 times larger than the amount of target information can be very difficult (\eg~a batch size of 10 in~\cite{zhu2019deep, zhao2020idlg}). 
When gradients are mixed with non-target information which is 30 times the amount of target information, the original information risk measured by sensitivity is extremely low (near zero). Also, extracting latent information from 30$\sim$50 times non-target information is hard (\eg~30$\sim$50 clients in~\cite{melis2019exploiting}).

\begin{figure*}[t]
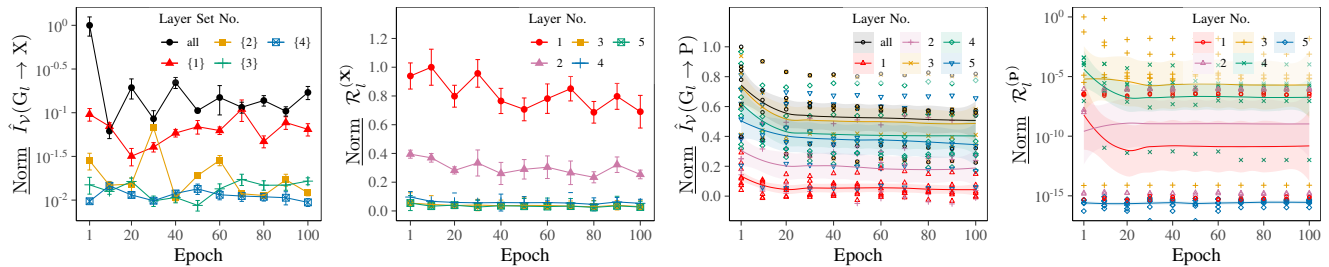

    \centering
    \begin{subfigure}{0.5\columnwidth}
        \iftex
            \resizebox{1\columnwidth}{!}{\input{figs/epoch_vinfo_x}}
        \else
            \includegraphics[width=1\columnwidth]{figs/epoch_vinfo_x.pdf}
        \fi
    \end{subfigure}%
    \begin{subfigure}{0.5\columnwidth}
        \iftex
            \resizebox{1\columnwidth}{!}{\input{figs/epoch_changes_sen}}
        \else
            \includegraphics[width=1\columnwidth]{figs/epoch_changes_sen.pdf}
        \fi
    \end{subfigure}%
    \begin{subfigure}{0.5\columnwidth}
        \iftex
            \resizebox{1\columnwidth}{!}{\input{figs/epoch_vinfo_y}}
        \else
            \includegraphics[width=1\columnwidth]{figs/epoch_vinfo_y.pdf}
        \fi
    \end{subfigure}%
    \begin{subfigure}{0.5\columnwidth}
        \iftex
            \resizebox{1\columnwidth}{!}{\input{figs/epoch_changes_sim}}
        \else
            \includegraphics[width=1\columnwidth]{figs/epoch_changes_sim.pdf}
        \fi
    \end{subfigure}
    \caption{Influence of training epoch on risks of original information [Left two] and latent information [Right two], measured on LeNet trained on CIFAR100 and LFW, respectively.}
    \label{fig:epoch}
    \vspace{-.3cm}
\end{figure*}

\mysubsubsection{Epochs.}
We measure the information leakage risks of gradients at specific epochs (from 1 to 100) during training. After epochs of training, the information leakage from gradients is calculated. 
The results show that overall the epoch only has a negligible impact on leakage risks (see Figure~\ref{fig:epoch}). Except for the leakage risk drop in the first several epochs (\ie~first 10 epochs for original information and first 20 epochs for latent information), epochs barely change gradients leakage risks. 
This may be due to that the magnitude of the updated gradients does not change significantly throughout these epochs (for fixed learning rate). In such a sense, in the first epochs, the machine learning optimizer quickly converges somewhere near the optimum and increases the model accuracy very quickly; in later epochs, the optimization becomes slower, and consequently, the magnitude of gradient changes stabilizes. 
We expect that with (much) more epochs of training, the information leakage may significantly change due to overfitting or convergence, but our analysis shows that with a reasonable number of epochs, it does not impact the leakage a lot.

\subsection{Impact of defense mechanisms}

\begin{figure}
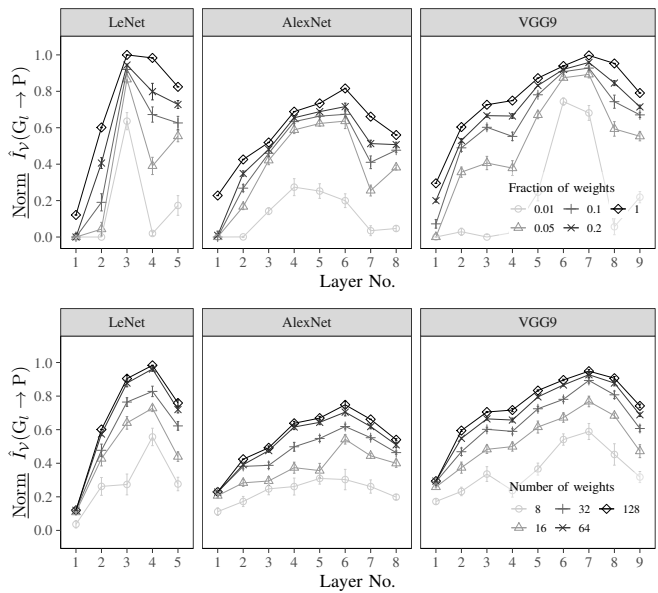

    \centering
    \iftex
        \resizebox{1\columnwidth}{!}{\input{figs/dropout_vinfo.tex}}
    \else
        \includegraphics[width=1\columnwidth]{figs/dropout_vinfo.pdf}
    \fi
    \caption{Usable information of the `race' attribute in a fixed number of gradients (Top) and a fraction of gradients (Bottom) from each layer of the neural network.}
    \label{fig:dropout_neurons}
    \vspace{-0.3cm}
\end{figure}

\mysubsubsection{Dropping fraction or number of gradients.}
In Figure~\ref{fig:dropout_neurons}, we present the usable information results when only parts of the gradients are available as side information. 
It is shown that with a particular dropping rate, the first layer of the classifier always has the highest usable information for the target property. Having only 5\% of the gradients (or even 1\% of the sensitive layers) available can still leak a great amount of private property information compared with having complete gradients.
However, the way of keeping a \emph{fixed fraction} of gradients can be influenced by the total number of gradients of the layer. We then measure the usable information from a \emph{fixed number} of gradients from each layer. Interestingly, the results still show that classifiers' first layers contain the most private latent information.

\begin{figure}[t!]
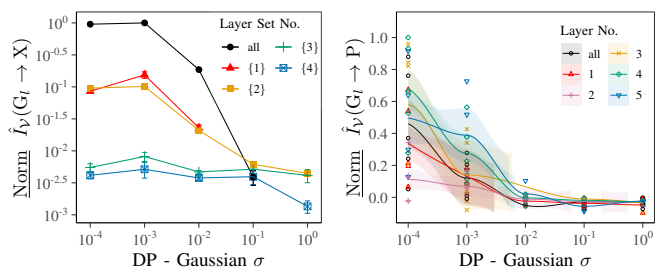

    \centering
    \begin{subfigure}{0.5\columnwidth}
        \iftex
            \resizebox{1\columnwidth}{!}{\input{figs/dp_vinfo_x}}
        \else
            \includegraphics[width=1\columnwidth]{figs/dp_vinfo_x.pdf}
        \fi
    \end{subfigure}%
    \begin{subfigure}{0.5\columnwidth}
        \iftex
            \resizebox{1\columnwidth}{!}{\input{figs/dp_vinfo_y}}
        \else
            \includegraphics[width=1\columnwidth]{figs/dp_vinfo_y.pdf}
        \fi
    \end{subfigure}%
    \caption{Influence of training epoch on risks of original information [Left] and latent information [Right], measured on LeNet trained on CIFAR100 and LFW, respectively.}
    \label{fig:dp}
\end{figure}

\mysubsubsection{Differential privacy.}
Adding DP noises before releasing gradients could be one way to reduce the privacy risks. Here we follow~\cite{shokri2015privacy, jayaraman2019evaluating}, i) to clip gradients using $l_2$-norm and then ii) to add Gaussian noises. 
Specifically, we set the max norm as $1$ for all cases and adjust the standard deviation $\sigma$ of Gaussian noises from $10^{0}$to $10^{-4}$. Noises are only added every time before releasing gradients; that is, pre-sample noise addition for DRAs, pre-batch (32 samples) noise addition for AIAs. We do not report the privacy budget because this is not a fully DP training (\ie~we do not test accuracy and focus on privacy analysis). 
Note that adding noises to gradients does not change the derivative results of $\frac{\partial \mathbf{g}_l(\bm{X})}{\partial \bm{X}}$, \ie~sensitivity. 
The empirical results (Figure~\ref{fig:dp}) show that adding noises to the first layers has higher effectiveness in alleviating original information risks, as the risks do not change significantly regarding the last layers. 
Similarly, noises could provide better protection in the last layers for latent information.
This again highlights the locations of different types of information and the need for potential layer-wise DP protection.

\section{Discussion}

\mysubsubsection{On the localization of the most sensitive layers.}
The location of the most sensitive layers in a model can change depending on the type of information and the attack family we consider. 
Our analysis showed that the nodes in the first layer or nodes in the middle layers of some deeper neural networks can contain the most sensitive information of the original input~\cite{geiping2020inverting, zhu2019deep}.
Regarding latent information, we found that the first FC layers (typically placed after the last Conv layer) are the most sensitive ones. 
One explanation for these observations is that latent information is more specific than the original information,~$\mathrm{X}$, but still more general than output information,~$\mathrm{Y}$, thus it may be best captured in layers between them. 
Specifically, discovering a latent attribute needs the extraction of high-level features; the FC layer takes the feature maps from the previous Conv layers, and classifies the input data by combining these discerning high-level features into a prediction for the latent attribute. Similarly, the last Conv layer is likely to have more latent information than previous Conv layers, because initial Conv layers learn generally learn more fine-grained information (\eg~ambient colors), while the latter Conv layers focus on high-level latent information (\eg~face identity); as also shown in~\cite{zeiler2014visualizing, mahendran2015understanding}.

\mysubsubsection{On the insights gained from gradient sensitivity.}
Sensitivity is a local measure that allows accounting for flatness in the objective function minimized in attacks.
This flatness is known to occur in neural network loss functions~\cite{draxler2018essentially}, and therefore the proposed sensitivity metric can be a valuable tool in understanding sensitive information leakage. 
In addition, a low sensitivity has been linked to increased robustness of trained models~\cite{sokolic2017robust}. 
Adapting the tools that were consequently introduced to improve robustness can help design better defenses against original information leakage. The gradient subspace distances are a novel metric introduced with the goal of capturing more \emph{coarse-grained} information. As sensitivity in itself is a local measure to capture latent information leakages, subspace distances are a valuable extension. 
Defined as the distance between gradients with and without target information, it helps inform in which trained models leaks can happen, or in other words whether the weights of the models are trained so they are able to distinguish between attributes. 

\mysubsubsection{On using localization for the design of protection mechanisms.}
Our characterization allows us to \emph{localize} which layers/nodes in which layers leak the most sensitive information. This highlights opportunities for designing defenses that are flexible and practical at a layer-wise level. 
For example, fully homomorphic encryption (FHE)-based approaches fully respect the model privacy but it leads to high runtime overhead. Our analysis can be used towards solutions utilizing FHE only on the sensitive part of the model during training. 
In addition, given the resource constraints on edge devices, TEE-based approaches with model partitioning have also become a promising approach. Specifically, during learning, one can deploy and run the most sensitive layers (\eg~the last Conv layer and the first FC layer) inside the TEE using model partitioned execution techniques across trusted and untrusted environments similar to~\cite{mo2020darknetz}. 
Our measures could also provide insights for layer-wise, federated training of models~\cite{mo2021ppfl, wang2019federated}, which could further reduce the privacy risk by exposing only specific layers instead of the complete model.

\mysubsubsection{On giving insights for defense mechanisms.} there are privacy-preserving techniques proposing to add noise to gradients (\ie~DP), share fewer gradients, or use dimensionality reduction and regularization (\eg~Dropout). 
However, none of them can protect against latent or original information leakage without significantly compromising model utility~\cite{melis2019exploiting, zhu2019deep}. 
The ability to localize the most sensitive layers can allow adding noise only where necessary. 
As one example, clipping or adding noises differently for layers in DP~\cite{mcmahan2018general, pichapati2019adaclip} and directly hiding layers in TEEs~\cite{mo2020darknetz} could be one promising protection without compromising the utility of the other layers and our quantification helps in understanding and designing these layer-wise protection mechanisms. 

\mysubsubsection{Future works}.
i) It is expected that DNNs with skip connections (\eg~ResNet) could give similar results because they usually have a unidirectional gradient flow, but this needs further experiments. Information leakages in other widely used networks such as Recurrent~\cite{sherstinsky2020fundamentals, wu2020comprehensive} or Graph neural networks need investigating;
ii) More defense mechanisms can be analyzed using our proposed metrics. For example, the influence of differential privacy can provide a further understanding of the linkage between information leakages and gradient changes;
iii) Factors other than sensitivity deserve further exploration. Bayesian-based generalization analysis~\cite{wilson2020bayesian} may also help to theoretically characterize information leakages.

\section{Conclusions}
Quantifying the information flow in backward propagation and information leakages associated with the computed gradients is still a conundrum, as it remains unclear in what part of the model, and for what kind of attacks, leakages happen. 
In this paper, we presented a framework that encompasses usable information on attack models and generalizes it by measuring the information loss from gradients over a certain \emph{family} of attack models. We also presented \textit{gradient-based} metrics. These metrics work directly on the trained model and are motivated by the mathematical formulation of successful attacks. Empirical results show that our layer-wise analysis provides a better understanding of the memorization of information in neural networks and facilitates the design of flexible layer-level defenses for establishing better trade-offs between privacy and costs. Specifically, we introduced mathematically-grounded tools to better quantify information leakages and applied these tools to localize sensitive information in several models over different datasets and training hyperparameters.

\bibliographystyle{IEEEtran}
\bibliography{references}
\balance

\end{document}